%% file: root.tex
\documentclass[journal]{IEEEtran}  % Comment this line out if you need a4paper

\IEEEoverridecommandlockouts                              % This command is only needed if 
\usepackage{graphicx}
\usepackage{amsmath} % assumes amsmath package installed
\usepackage{mathtools}
\usepackage{amssymb}  % assumes amsmath package installed
\usepackage{stﬂoats}

\usepackage{subcaption}
\usepackage{array}
\usepackage{flushend}
\usepackage{cite}   % collect references numbers inside single set of square brackets

\input{notation}
\usepackage{xcolor}
\usepackage{hyperref}

\renewcommand{\q}{\vec{q}}
\newcommand{\w}{\boldsymbol{\omega}}

\newcommand{\qd}{\dot{\vec{q}}}
\newcommand{\qdd}{\ddot{\vec{q}}}
\newcommand{\J}{\mat{J}}

\newcommand{\Jt}{\mat{J_{\!t}}}
\newcommand{\JR}{\mat{J_{\!R}}}
\renewcommand{\R}{\mat{R}}
\newcommand{\Rd}{\dot{\mat{R}}}
\newcommand{\Rdd}{\ddot{\mat{R}}}
\newcommand{\T}{\mat{T}}
\newcommand{\Td}{\dot{\mat{T}}}
\newcommand{\Tdd}{\ddot{\mat{T}}}
\renewcommand{\t}{\vec{t}}
\newcommand{\td}{\dot{\vec{t}}}
\newcommand{\tdd}{\ddot{\vec{t}}}

\newcommand{\Jv}{\mat{J}_v}
\newcommand{\Jw}{\mat{J_{\!\omega}}}

\renewcommand{\H}{\mat{H}}

\newcommand{\Ht}{\mat{H_{\!t}}}
\newcommand{\HR}{\mat{H_{\!R}}}
\newcommand{\Hw}{\mat{H}_{\!\alpha}}
\newcommand{\Hwx}[1]{\mat{H}_{{\!\alpha}_{#1}}}
\newcommand{\Hv}{\mat{H}_{\!a}}
\newcommand{\Hvx}[1]{\mat{H}_{{\!a}_{#1}}}

\newcommand{\TR}[1]{\mat{T}_{{\!\R}_{#1}}}
\newcommand{\Tt}[1]{\mat{T}_{{\!\t}_{#1}}}

\makeatletter
\newcommand*{\balancecolsandclearpage}{%
  \close@column@grid
  \clearpage
  \twocolumngrid
}
\makeatother

\title{\LARGE \bf
A Systematic Approach to Computing the Manipulator Jacobian and Hessian using the Elementary Transform Sequence}

\author{Jesse Haviland, Peter Corke, \IEEEmembership{Fellow,~IEEE}% <-this % stops a space
\thanks{Manuscript received xxxxxxx xx, 20XX; revised xxxxxx xx, 20XX. This research was conducted by the Australian Research Council project number CE140100016, and supported by the QUT Centre for Robotics.}
\thanks{The authors are with the Australian Centre for Robotic Vision (ACRV), Queensland University of Technology (QUT), Brisbane, Australia.}
}

\IEEEpubid{0000--0000/00\$00.00~\copyright~20xx xxxx}

\begin{document}

\maketitle
% \thispagestyle{empty}
% \pagestyle{empty}

%%%%%%%%%%%%%%%%%%%%%%%%%%%%%%%%%%%%%%%%%%%%%%%%%%%%%%%%%%%%%%%%%%%%%%%%%%%%%%%%
\begin{abstract}

The elementary transform sequence (ETS) provides a universal method of describing the kinematics of any serial-link manipulator. 
The ETS notation is intuitive and easy to understand, 
while avoiding the complexity and limitations of Denvit-Hartenberg frame assignment.
In this paper, we describe a systematic method for computing the manipulator Jacobian and Hessian (differential kinematics) using the ETS notation. 
Differential kinematics have many applications including numerical inverse kinematics, resolved-rate motion control and manipulability motion control.
Furthermore, we provide an open-source Python library which implements our algorithm and can be interfaced with any serial-link manipulator (available at
% github.com/petercorke/robotics-toolbox-python).
\href{https://github.com/petercorke/robotics-toolbox-python}{github.com/petercorke/robotics-toolbox-python}).

\end{abstract}


%%%%%%%%%%%%%%%%%%%%%%%%%%%%%%%%%%%%%%%%%%%%%%%%%%%%%%%%%%%%%%%%%%%%%%%%%%%%%%%%

\section{Introduction}
\input{1_introduction.tex}

\section{The Elementary Transform Sequence}\label{sec:ets}

\input{2_ets.tex}

\section{First Derivative of a Pose} \label{sec:dT}
\input{4_dT.tex}

\section{First Derivative of an Elementary Transform} \label{sec:dE}
\input{5_dE.tex}

\section{The Manipulator Jacobian} \label{sec:dETS}
\input{6_dETS.tex}

\section{Second Derivative of a Pose} \label{sec:ddETS}
\input{7_ddT.tex}

\section{Second Derivative of an Elementary Transform} \label{sec:ddE}
\input{8_ddE.tex}

\section{The Manipulator Hessian} \label{sec:ddH}

\input{9_ddETS.tex}

\appendices
\input{9_app.tex}

\bibliographystyle{IEEEtran} 
\bibliography{ref} % For overleaf

\end{document}

%% file: notation.tex
\usepackage{amsmath, amssymb}

\usepackage{xifthen}
\usepackage{color}

\newcommand{\ba}{\begin{eqnarray}}
\newcommand{\ea}{\end{eqnarray}}
\newcommand{\eq}[1]{(\ref{#1})}

\newcommand{\R}{\mathbb{R}}

\newcommand{\presup}[1]{\,{}^{\scriptscriptstyle #1}\!}

\newcommand{\pose}[1][ZZZZ]{\ifthenelse{\equal{#1}{ZZZZ}}{}{\presup{#1}}{\mathbf{\xi}}}
\newcommand{\estpose}[1][ZZZZ]{\ifthenelse{\equal{#1}{ZZZZ}}{}{\presup{#1}}{\mathbf{\hat{\xi}}}}
\newcommand{\hpose}[1][ZZZZ]{\ifthenelse{\equal{#1}{ZZZZ}}{}{\presup{#1}}{\hat{\mathbf{\xi}}}}
\newcommand{\posedot}[1][ZZZZ]{\ifthenelse{\equal{#1}{ZZZZ}}{}{\presup{#1}}{\mathbf{\nu}}}
\newcommand{\q}[1][ZZZZ]{\ifthenelse{\equal{#1}{ZZZZ}}{}{\presup{#1}}{\mathring{q}}}
\DeclareMathAlphabet{\mathitbf}{OML}{cmm}{b}{it}
\newcommand{\twist}[2][ZZZZ]{\ifthenelse{\equal{#1}{ZZZZ}}{}{\presup{#1}}{\mathcal{S}}}
\renewcommand{\vec}[2][ZZZZ]{\ifthenelse{\equal{#1}{ZZZZ}}{}{\presup{#1}}{\mathitbf{#2}}}
\newcommand{\hvec}[2][ZZZZ]{\ifthenelse{\equal{#1}{ZZZZ}}{}{\presup{#1}}{\tilde{\vec{#2}}}}
\newcommand{\evec}[2][ZZZZ]{\ifthenelse{\equal{#1}{ZZZZ}}{}{\presup{#1}}{\hat{\vec{#2}}}}
\newcommand{\bvec}[2][ZZZZ]{\ifthenelse{\equal{#1}{ZZZZ}}{}{\presup{#1}}{\bar{\vec{#2}}}}
\newcommand{\dvec}[2][ZZZZ]{\ifthenelse{\equal{#1}{ZZZZ}}{}{\presup{#1}}{\dot{\vec{#2}}}}
\newcommand{\ddvec}[2][ZZZZ]{\ifthenelse{\equal{#1}{ZZZZ}}{}{\presup{#1}}{\ddot{\vec{#2}}}}

\newcommand{\mat}[2][ZZZZ]{\ifthenelse{\equal{#1}{ZZZZ}}{}{\presup{#1}\,}{{\boldsymbol #2}}}
\newcommand{\dmat}[2][ZZZZ]{\ifthenelse{\equal{#1}{ZZZZ}}{}{\presup{#1}\,}{{\dot{\boldsymbol #2}}}}
\newcommand{\emat}[2][ZZZZ]{\ifthenelse{\equal{#1}{ZZZZ}}{}{\presup{#1}\,}{\hat{\boldsymbol#2}}}
\newcommand{\matfn}[3][ZZZZ]{\ifthenelse{\equal{#1}{ZZZZ}}{}{\presup{#1}}{{\mat{#2}}\left(#3\right)}}
\newcommand{\Rt}[2][ZZZZ]{\ifthenelse{\equal{#1}{ZZZZ}}{}{\presup{#1}}{{\bf R}\left(#2\right)}}

\newcommand{\point}[2][ZZZZ]{\ifthenelse{\equal{#1}{ZZZZ}}{}{\presup{#1}}{\mathbf{\mathrm{#2}}}}

\newfont{\School}{pncr}
\newfont{\eightTR}{pncr at 8pt}
\usepackage{color}
\usepackage{fancyvrb}
\fvset{formatcom=\color{blue},fontseries=c,fontfamily=courier,xleftmargin=4mm,commentchar=!}
\DefineVerbatimEnvironment{Code}{Verbatim}{formatcom=\color{blue},fontseries=c,fontfamily=courier,fontsize=\footnotesize,xleftmargin=4mm,commentchar=!}
\DefineVerbatimEnvironment{CodeSmall}{Verbatim}{formatcom=\color{blue},fontseries=c,fontfamily=courier,fontsize=\scriptsize,xleftmargin=1mm,commentchar=!}
\DefineVerbatimEnvironment{CodeNum}{Verbatim}{numbers=left,numbersep=4pt,formatcom=\color{blue},fontseries=c,fontfamily=courier,fontsize=\footnotesize,xleftmargin=4mm}
\newcommand{\model}[1]{\index{code}{#1@\textit{#1}}\ifthenelse{\boolean{draft}}{{\color{green}\Verb+#1+}}{\Verb+#1+}}
\newcommand{\block}[1]{\ifthenelse{\boolean{draft}}{{\color{green}\Verb+#1+}}{\textsf{#1}}}
\newcommand{\func}[2][ZZZZ]{\ifthenelse{\equal{#1}{ZZZZ}}{\index{code}{#2}}{\index{code}{#1}}\ifthenelse{\boolean{draft}}{{\color{green}\Verb+#2+}}{\Verb+#2+}}
\newcommand{\methodb}[2]{\index{code}{#1@\textbf{#1}!.#2}\ifthenelse{\boolean{draft}}{{\color{magenta}\Verb+#1.#2+}}{\Verb+#1.#2+}}
\newcommand{\method}[2]{\index{code}{#1@\textbf{#1}!.#2}\ifthenelse{\boolean{draft}}{{\color{magenta}\Verb+#2+}}{\Verb+#2+}}
\newcommand{\class}[1]{\index{code}{#1@\textbf{#1}}\ifthenelse{\boolean{draft}}{{\color{cyan}\Verb+#1+}}{\Verb+#1+}}
\newcommand{\property}[1]{\index{property}{#1}\ifthenelse{\boolean{draft}}{{\color{cyan}\Verb+#1+}}{\Verb+#1+}}

\newcommand{\SE}[1]{\ensuremath{\mathrm{{\bf SE}(#1)}}}
\newcommand{\SO}[1]{\ensuremath{\mathrm{{\bf SO}(#1)}}}
\newcommand{\se}[1]{\ensuremath{\mathrm{{\bf se}(#1)}}}
\newcommand{\so}[1]{\ensuremath{\mathrm{{\bf so}(#1)}}}
\newcommand{\isk}[1]{\vee\left( #1\right)}
\newcommand{\iskx}[1]{\vee_{\times}\left( #1\right)}
\newcommand{\skx}[1]{\left[#1\right]_{\times}}
\newcommand{\sk}[1]{\left[#1\right]}

%% file: 1_introduction.tex
\begin{figure}[!t]
    \centering
    \includegraphics[height=13.4cm]{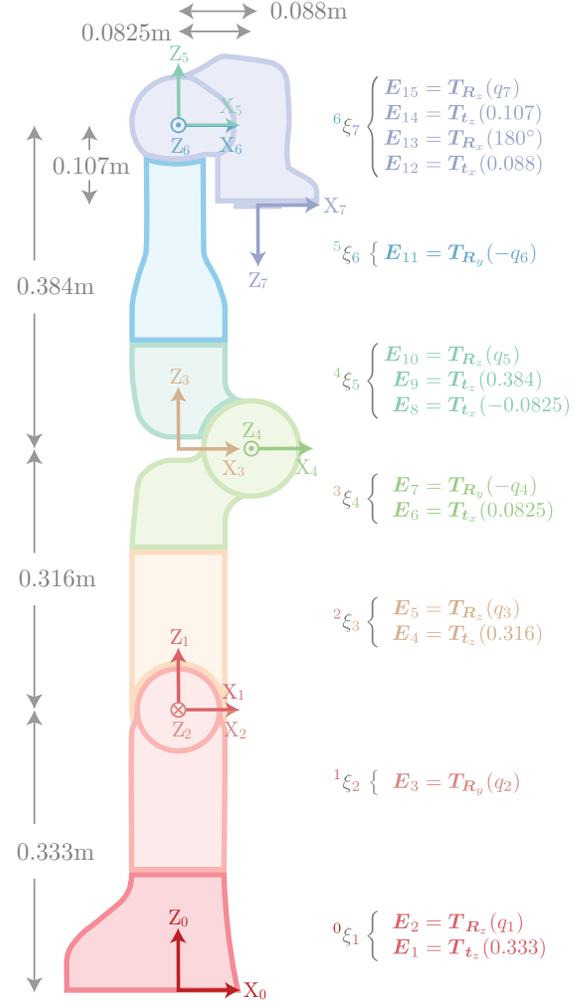} %14.1
    \caption{
        The Elementary Transform Sequence of the 7 degree-of-freedom Franka-Emika Panda serial-link manipulator in its zero-angle configurations. $\mat{E}_i$ represents an elementary transform while $^a\pose_b$ 
        represents the pose of link frame $b$ in the reference frame of link $a$.
    }
    \vspace{-0.6cm}
    \label{fig:cover}
\end{figure}

Robot kinematics is an essential area of study and provides the foundation for robotic control. The elementary transform sequence (ETS), introduced in \cite{ets}, provides a universal method for describing the kinematics of a serial-link manipulator. This intuitive and systematic approach can be calculated with a simple \emph{walk through} procedure. The resulting sequence comprises a number of elementary translations and rotations, from the base frame of the robot to the end-effector. An example of an ETS is displayed in Figure \ref{fig:cover} for the Franka-Emika Panda in its zero-angle configuration.

The ETS is conceptually easy to grasp, 
since it avoids the frame assignment constraints
of Denavit and Hartenberg (DH) notation \cite{dh}, and allows 
joint rotation or translation about or along any axis. However, 
DH notation underpins a large body of algorithms 
for standard applications
such as kinematics, differential kinematics, and dynamics of a robot with applications to motion control, planning and simulation \cite{peter, seth}.

The manipulator Jacobian is a fundamental tool for robotic control and has many applications, including inverse kinematics \cite{ik1, ik2}, motion planning \cite{pp1, pp2}, and resolved-rate motion control \cite{rrmc}. The manipulator Hessian has applications in more advanced controllers \cite{hess}, typically where the derivative of the Jacobian is used, such as manipulability motion control \cite{mmc, mmc2, neo}, high speed robotics \cite{hs1}, or dynamics \cite{d1, d2}.

The work in \cite{ets} goes on to describe an algebraic procedure that can be automatically applied to an ETS sequence to factorize it into the standard or modified Denavit-Hartenberg form. This allows an ETS model to access the large body of standard equations such as in \cite{peter}. However, this work has a different philosophy; in order to take advantage of the universal nature of the ETS, all kinematic algorithms should be based directly on ETS, rather than DH, notation. 

\enlargethispage{-1.3\baselineskip}

In this paper, we describe a systematic approach to calculating the forward kinematics, manipulator Jacobian and the manipulator Hessian directly from the ETS representation. Furthermore, we provide an open source Python library which implements our approach along with a DH to ETS converter to provide maximum usability for robots irrespective of its representation.

%% file: 2_ets.tex
The elementary transform sequence (ETS) is an intuitive and systematic approach to describing the kinematic model of a serial-link manipulator \cite{peter}. The ETS is a string of elementary translations and rotations, from the user defined base coordinate frame to the end-effector frame of the robot.

The forward kinematics of a serial-link manipulator provides a non-linear mapping

\begin{equation*}
    \T(t) = {\cal K}(\q(t))
\end{equation*}
between the joint space and Cartesian task space,
where $\q(t) = (q_1(t), q_2(t), \cdots q_n(t)) \in \mathbb{R}^n$ is the vector of joint generalized coordinates, $n$ is the number of joints, and $\T \in \SE{3}$ is a homogeneous transformation matrix representing the pose of the robot's end-effector in the world-coordinate frame. The ETS model defines $\cal{K}(\cdot)$ as the product of $M$ elementary transforms $\mat{E}_i \in \SE{3}$

\begin{align} \label{m:ets1}
    \T(t) &= \mat{E}_1(\eta_1) \mat{E}_2(\eta_2) \hdots \mat{E}_M(\eta_M) \nonumber\\
    &= \prod_{i=1}^{M} \mat{E}_i(\eta_i).
\end{align}

\begin{figure}[!t]
    \centering
    \includegraphics[height=15cm]{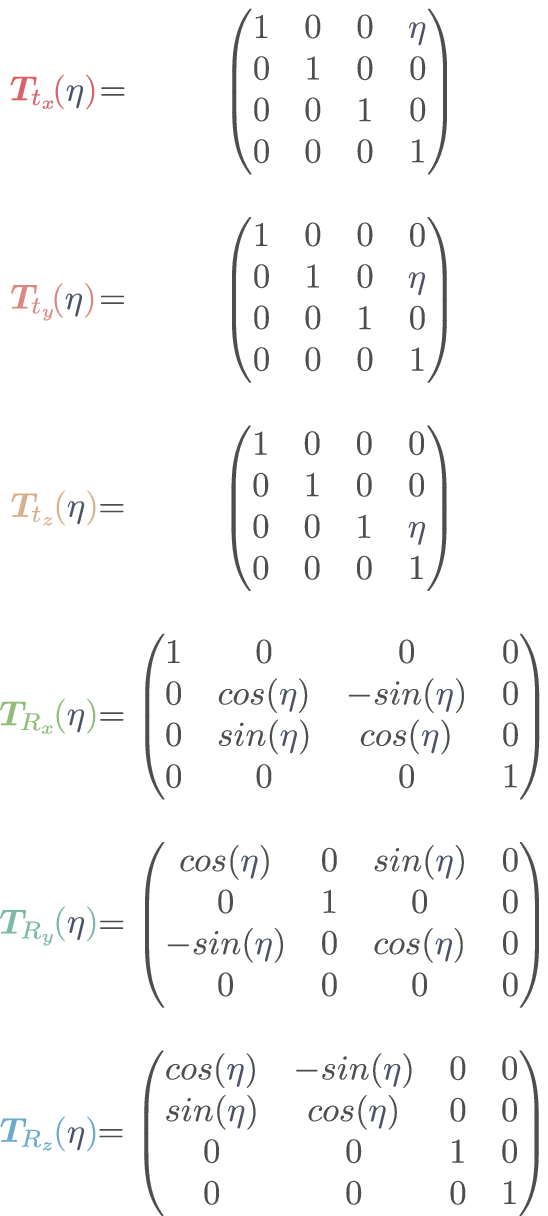}
    \caption{
        The six different elementary transforms $\mat{E} \in \SE{3}$ from (\ref{eq:Ei}) which are the building blocks for ETS notation. Each homogeneous transformation matrix above represents a translation along or a rotation about a single axis which is paramaterized by $\eta$ as defined in (\ref{eq:eta1}) and (\ref{eq:eta2})
    }
    \label{fig:et}
\end{figure}

Each of the elementary transforms can be a pure translation along, or a pure rotation about the x, y, or z-axis by an amount $\eta_i$. Explicitly, each transform is one of the following

\begin{equation} \label{eq:Ei}
    \mat{E}_i = 
    \left\{
    \begin{matrix}
        \Tt{x}(\eta_i) \\
        \Tt{y}(\eta_i) \\
        \Tt{z}(\eta_i) \\
        \TR{x}(\eta_i) \\
        \TR{y}(\eta_i) \\
        \TR{z}(\eta_i) \\
    \end{matrix}
    \right.
\end{equation}
where each of the matrices are displayed in Figure \ref{fig:et} and the parameter $\eta_i$ is either a constant $c_i$ (translational offset or rotation) or a joint variable $q_j(t)$

\begin{equation} \label{eq:eta1}
    \eta_i = 
    \left\{
    \begin{matrix}
        c_i \\
        q_j(t) \\
    \end{matrix}
    \right. 
\end{equation}
and the joint variable is

\begin{equation} \label{eq:eta2}
    q_j(t) = 
    \left\{
    \begin{matrix*}[l]
        \theta(t) & \quad \mbox{for a revolute joint}\\
        d(t) & \quad \mbox{for a prismatic joint}\\
    \end{matrix*}
    \right. 
\end{equation}
where $\theta(t)$ represents a joint angle, and $d(t)$ represents a joint translation.

As shown in Figure \ref{fig:cover}, we can also calulcate the pose of a link frame $b$ relative to a previous link frame $a$ as $^a\pose_b$. Calculating this is trivial and uses a subset of the ETS

\begin{align} \label{m:joint_pose}
    ^a\pose_b &= \prod_{i=a}^{b} \mat{E}_i(\eta_i).
\end{align}

%% file: 4_dT.tex
Now consider the end-effector pose, which varies as a function of joint coordinates. The derivative with respect to time is

\begin{equation} \label{eq:Tdot}
\Td = 
    \frac{\mathrm{d} \mat{T}}
         {\mathrm{d}t} 
    = 
    \frac{\partial \T}
         {\partial q_1} \dot{q}_1 + \cdots +  
    \frac{\partial \T}
         {\partial q_n} \dot{q}_n \in \mathbb{R}^{4 \times 4} 
\end{equation}
where each $\frac{\partial \T}{\partial q_i} \in \mathbb{R}^{4 \times 4}$.

% As shown in Figure \ref{fig:transform}, 
The information in $\T$ is non-minimal, and redundant, as is the information in $\Td$.  We can write these respectively as

\begin{equation}
     \T = 
     \begin{pmatrix}
          \R & \t \\ 
          0 & 1
     \end{pmatrix}
     , \,\,\, \Td = 
     \begin{pmatrix}
          \Rd & \td \\ 
          0 & 0 
     \end{pmatrix}
\end{equation}
where $\R \in \SO{3}$ and $\t \in \mathbb{R}^3$.

We will write the partial derivative in partitioned form as

\begin{equation} \label{eq:partition}
     \frac{\partial \T}
          {\partial q_i} = 
     \begin{pmatrix}
          \JR_i & \Jt_i \\ 
          0 & 0 
     \end{pmatrix} 
\end{equation}
where $\JR_i \in \mathbb{R}^{3 \times 3}$ and $\Jt_i \in \mathbb{R}^{3 \times 1}$, and then rewrite \eq{eq:Tdot} as 

\begin{equation*}
     \begin{pmatrix}
          \Rd & \td \\ 0 & 0 
     \end{pmatrix} = 
     \begin{pmatrix} 
          \JR_1 & \Jt_1 \\ 0 & 0 
     \end{pmatrix}
     \dot{q}_1 + \cdots +   
     \begin{pmatrix} 
          \JR_n & \Jt_n \\ 0 & 0 
     \end{pmatrix}
     \dot{q}_n \,\,.
\end{equation*}
and write a matrix equation for each non-zero partition

\begin{align}
     \Rd &= \JR_1  \dot{q}_1 + \cdots +   \JR_n  \dot{q}_n \label{eq:Rdot}\\
     \td &= \Jt_1  \dot{q}_1 + \cdots +   \Jt_n  \dot{q}_n \label{eq:tdot}
\end{align}
where each term represents the contribution to end-effector velocity due to motion of the corresponding joint.

Taking (\ref{eq:tdot}) first, we can simply write

\begin{align}
     \td &= 
     \begin{pmatrix}
          \Jt_1 & \cdots  & \Jt_n 
     \end{pmatrix} 
     \begin{pmatrix}
          \dot{q}_1 \\ \vdots \\  \dot{q}_n 
     \end{pmatrix} \nonumber \\
     &= 
     \Jv(\q) \qd  \label{eq:td}
\end{align}
where $\Jv(\q) \in \mathbb{R}^{3 \times n}$ is the translational part of the manipulator Jacobian.

Rotation rate is slightly more complex, but using the identity $\Rd = \skx{\w} \R$ where $\skx{\w} \in \so{3}$ is a skew-symmetric matrix, we can rewrite (\ref{eq:Rdot}) as

\begin{equation} \label{eq:skxr}
     \skx{\w} \R = \JR_1  \dot{q}_1 + \cdots +   \JR_n  \dot{q}_n 
\end{equation}
and rearrange to 

\begin{figure*}[!t]
     \centering
     \includegraphics[height=4.5cm]{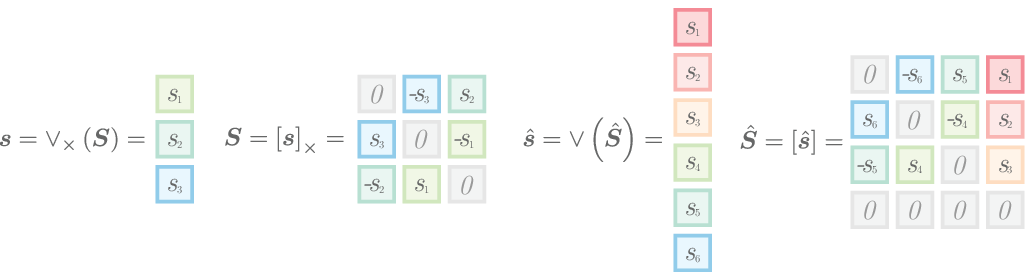}
     \caption{
          Shown on the left is a vector $\vec{s}\in \mathbb{R}^3$ along with its corresponding skew symmetric matrix $\mat{S} \in \mathbb{R}^{3 \times 3}$. Shown on the right is a vector $\evec{s}\in \mathbb{R}^6$ along with its corresponding augmented skew symmetric matrix $\emat{S} \in \mathbb{R}^{4 \times 4}$. The skew functions $\skx{\cdot} : \mathbb{R}^3 \mapsto \mathbb{R}^{3\times3}$ maps a vector to a skew symmetric matrix, and $\sk{\cdot} : \mathbb{R}^6 \mapsto \mathbb{R}^{4\times4}$ maps a vector to an augmented skew symmetric matrix. The inverse skew functions $\iskx{\cdot} : \mathbb{R}^{3\times3} \mapsto \mathbb{R}^3$ maps a skew symmetric matrix to a vector and $\isk{\cdot} : \mathbb{R}^{4\times4} \mapsto \mathbb{R}^6$ maps an augmented skew symmetric matrix to a vector.
     }
     \label{fig:skew}
 \end{figure*}

\begin{equation*}
     \skx{\w}  = (\JR_1 \R^\top) \dot{q}_1 + \cdots +   (\JR_n \R^\top) \dot{q}_n  \in \so{3}
\end{equation*}
where each of the terms $\JR_i \R^\top$ must also be skew-symmetric since the sum of skew-symmetric matrices equals a skew-symmetric matrix. % since from \eq{eq:partition} we have $\JR = \Rd$.  
This $3\times 3$ matrix equation therefore has only 3 unique equations so applying the inverse skew operator described in Figure \ref{fig:skew} to both sides we have

\begin{align} \label{eq:Rd}
     \vec{\omega}  
     &= 
     \iskx{\JR_1 \R^\top } \dot{q}_1 + \cdots + 
     \iskx{\JR_n \R^\top} \dot{q}_n \nonumber \\
     &= 
     \bigg(
     \begin{matrix}
           \iskx{\JR_1 \R^\top} &  \cdots  &  \iskx{\JR_n \R^\top}
     \end{matrix}
     \bigg)
     \begin{pmatrix}
          \dot{q}_1 \\ 
          \vdots \\  
          \dot{q}_n 
     \end{pmatrix} \nonumber \\
     &= \Jw(\q) \qd 
\end{align}
where $\Jw(\q)  \in \mathbb{R}^{3 \times n}$ is the rotational part of the manipulator Jacobian.

Combining \eq{eq:td} and \eq{eq:Rd} we can write

\begin{equation} \label{eq:jacob}
     \begin{pmatrix} 
          \vec{v} \\ 
          \w 
     \end{pmatrix} = 
     \begin{pmatrix} 
          \Jv(\q) \\ 
          \Jw(\q) 
     \end{pmatrix} \qd  
\end{equation}
which is the time derivative of the manipulator forward kinematics and

\begin{equation} \label{eq:JT}
     \J(\q) = 
     \begin{pmatrix} 
          \Jv(\q) \\ 
          \Jw(\q) 
     \end{pmatrix} \in \mathbb{R}^{6\times n}
\end{equation}
is the manipulator Jacobian matrix expressed in the world-coordinate frame. More compactly we write

\begin{equation} \label{eq:j0}
     \begin{pmatrix} 
          \vec{v} \\ 
          \w 
     \end{pmatrix}
     = \J(\q) \qd.
\end{equation}
which provides the derivative of the left side of (\ref{m:ets1}). However, to actually compute (\ref{eq:j0}), we need to first find the derivative of a pose with respect to a joint coordinate.

%% file: 5_dE.tex
Before differentiating the ETS, it is useful to consider the derivative of a single Elementary Transform. An overview of \SO{3}, \SE{3}, Lie groups, Lie algebras and skew-symmetric matrices
have been provided in Appendix \ref{sec:lie} and \ref{sec:lie2}.
\\

\paragraph{Derivative of a Rotation}

Consider a rotation in exponential form

\begin{equation*}
    \mat{R}(\theta(t)) = \textrm{e}^{ \skx{\evec{\omega}} \theta(t)} \in \SO{3}
\end{equation*}
where the rotation is described by the rotation axis which is the unit vector $\evec{\omega}$, and a rotation angle $\theta(t)$. The derivative of a rotation with respect to the rotation angle is 

\begin{align} \label{eq:dr1}
    \frac{\mathrm{d}\mat{R}(\theta)}
         {\mathrm{d} \theta} 
    &=
    \skx{\evec{\omega}} \ \textrm{e}^{\skx{\evec{\omega}} \theta(t)} \nonumber \\
    &= \skx{\evec{\omega}} \mat{R}(\theta(t)) 
\end{align}
where multiplyling each side by $\frac{\mathrm{d} \theta}{\mathrm{d} t}$ gives the identity $\Rd = \skx{\w} \R$. Therefore, the following relationship holds

\begin{align*}
    \skx{\evec{\omega}} \dot{\theta} &= \skx{\vec{\omega}} \\
    \evec{\omega} \ \dot{\theta} &= \vec{\omega}
\end{align*}
and $\vec{\omega} \in \mathbb{R}^3$ is the angular velocity.

The angular velocity describes the instantaneous rate of rotation about the x, y, and z-axis. The rotation axis $\evec{\omega}$
can be recovered from (\ref{eq:dr1}) 
using the inverse skew operator

\begin{align}
    \evec{\omega} &= \iskx{ 
        \frac{\mathrm{d} \matfn{R}{\theta(t)}}
             {\mathrm{d} \theta}
    \matfn{R}{\theta(t)}^\top }  \label{m:dw1} 
\end{align}
since $ \mat{R} \in \SO{3}$, then $\mat{R}^{-1} = \mat{R}^\top$.

For an ETS, we only need to consider the elementary rotations $\mat{R}_x$, $\mat{R}_y$, and $\mat{R}_z$. These are embedded within \SE{3}, as $\TR{x}$, $\TR{y}$, and $\TR{z}$ which are pure rotations with no translational component. We can show that the derivative of each elementary rotation with respect to a rotation angle is

\begin{align}
    \dfrac{\mathrm{d} \TR{x}(\theta)}
          {\mathrm{d} \theta}
    &=     
    \begin{pmatrix}
        0 & 0 & 0 & 0 \\
        0 & 0 & -1 & 0 \\
        0 & 1 & 0 & 0 \\
        0 & 0 & 0 & 0 
    \end{pmatrix} \TR{x}(\theta) = \sk{\evec{s}_{\R_x}} \TR{x}(\theta), \label{eq:ERx} \\
    \dfrac{\mathrm{d} \TR{y}(\theta)}
          {\mathrm{d} \theta}
    &=     
    \begin{pmatrix}
        0 & 0 & 1 & 0 \\
        0 & 0 & 0 & 0 \\
        -1 & 0 & 0 & 0 \\
        0 & 0 & 0 & 0 
    \end{pmatrix} \TR{y}(\theta) = \sk{\evec{s}_{\R_y}} \TR{y}(\theta), \label{eq:ERy} \\
    \dfrac{\mathrm{d} \TR{z}(\theta)}
          {\mathrm{d} \theta}
    &=     
    \begin{pmatrix}
        0 & -1 & 0 & 0 \\
        1 & 0 & 0 & 0 \\
        0 & 0 & 0 & 0 \\
        0 & 0 & 0 & 0  
    \end{pmatrix} \TR{z}(\theta) = \sk{\evec{s}_{\R_z}} \TR{z}(\theta), \label{eq:ERz}
\end{align}
where each of the augmented skew symmetric matrices $\sk{\evec{s}_{\R}}$ above corresponds to one of the generators of \SE{3} which lies in \se{3}, the tangent space of \SE{3}. If a defined joint rotation is negative about the axis, as is $\mat{E}_7$ and $\mat{E}_{11}$ in the ETS of the Panda shown in Figure \ref{fig:cover}, then the negative of $\sk{\evec{s}_{\R}}$ is used to calculate the derivative.

\begin{figure}[!t]
    \centering
    \includegraphics[height=2.4cm]{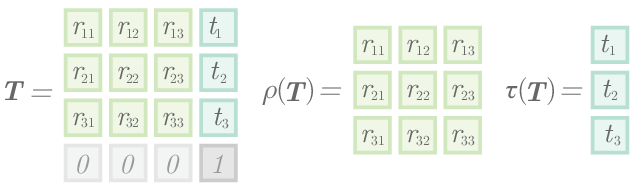}
    \caption{
        Visualization of a homogeneous transformation matrix (the derivatives share the form of $\mat{T}$, except will have a 0 instead of a 1 located at $\mat{T}_{44}$). Where the matrix $\rho(\mat{T}) \in \mathbb{R}^{3 \times 3}$ of green boxes forms the rotation component, and the vector $\tau(\mat{T}) \in \mathbb{R}^{3}$ of blue boxes form the translation component. The rotation component can be extracted throught the function $\rho(\cdot) : \mathbb{R}^{4\times 4} \mapsto  \mathbb{R}^{3\times 3}$, while the translation component can be extracted through the function $\tau(\cdot) : \mathbb{R}^{4\times 4} \mapsto  \mathbb{R}^{3}$.
    }
    \vspace{-0.5cm}
    \label{fig:transform}
\end{figure}

Equation (\ref{m:dw1})
only uses the rotational component of the pose. Using the function $\rho(\cdot)$ described in Figure \ref{fig:transform}
we can restate (\ref{m:dw1}) 
as 

\begin{align} \label{eq:raxis1}
    \evec{\omega} &= 
    \iskx{ 
        \rho
        \left(
            \frac{\mathrm{d} \TR{}(\theta(t))}
                 {\mathrm{d} \theta}
        \right)
        \rho
        \left(
            \TR{}(\theta(t))^\top 
        \right)
    }
\end{align}

Since $\R \R^\top = \mat{I}$, where $\mat{I}$ is the identity matrix, we can see that (\ref{eq:raxis1}) will lead to

\begin{equation}
    \evec{\omega}_x =
    \begin{pmatrix}
        1 \\ 0 \\ 0
    \end{pmatrix}, \quad 
    \evec{\omega}_y =
    \begin{pmatrix}
        0 \\ 1 \\ 0
    \end{pmatrix}, \quad 
    \evec{\omega}_z =
    \begin{pmatrix}
        0 \\ 0 \\ 1
    \end{pmatrix}, 
\end{equation}
which is quite intuitive as this result status that a revolute joint operating around an axis will only cause a velocity about that axis.

\paragraph{Derivative of a Translation}

Consider the three elementary translations $\Tt{}$ shown in Figure \ref{fig:transform}. Using the $\tau(\cdot)$ function, we can recover the pure translations $\vec{t}(d(t)) \in \mathbb{R}^3$ where $d(t)$ describes the offset distance.

A derivative of a translation is required when considering a prismatic joint. For an ETS, these translations are embedded in \SE{3} as $\Tt{x}$, $\Tt{y}$, and $\Tt{z}$ which are pure translations with no rotational component. We can show that the partial derivative of each elementary translation with respect to a translation offset $d(t)$ is 

\begin{align}
    \dfrac{\mathrm{d} \Tt{x}(d)}
          {\mathrm{d} d}
    &=     
    \begin{pmatrix}
        0 & 0 & 0 & 1 \\
        0 & 0 & 0 & 0 \\
        0 & 0 & 0 & 0 \\
        0 & 0 & 0 & 0 
    \end{pmatrix} = \sk{\evec{s}_{\vec{t}_x}}, \label{eq:ETx}\\
    \dfrac{\mathrm{d} \Tt{y}(d)}
          {\mathrm{d} d}
    &=     
    \begin{pmatrix}
        0 & 0 & 0 & 0 \\
        0 & 0 & 0 & 1 \\
        0 & 0 & 0 & 0 \\
        0 & 0 & 0 & 0 
    \end{pmatrix} = \sk{\evec{s}_{\vec{t}_y}}, \label{eq:ETy}\\
    \dfrac{\mathrm{d} \Tt{z}(d)}
          {\mathrm{d} d}
    &=     
    \begin{pmatrix}
        0 & 0 & 0 & 0 \\
        0 & 0 & 0 & 0 \\
        0 & 0 & 0 & 1 \\
        0 & 0 & 0 & 0 
    \end{pmatrix} = \sk{\evec{s}_{\vec{t}_z}}, \label{eq:ETz}
\end{align}
where each of the augmented skew symmetric matrices $\sk{\evec{s}_{\vec{t}}}$ above are the other three generators of \SE{3} which lie in \se{3}. If the translation is negative along an axis, then the negative of $\sk{\evec{s}_{\vec{t}}}$ should be used to calculate the derivative.

Using the function $\tau(\cdot)$ described in Figure \ref{fig:transform},
which maps the translational component of a $4 \times 4$ matrix to a $3$ vector, we can write the translation axis as 

\begin{align} \label{eq:dt1}
    \evec{v}
    &=
    \tau 
    \left(
        \dfrac{\mathrm{d} \Tt{}(d)}
              {\mathrm{d} d}
    \right)
\end{align}
where 

\begin{align}
    \vec{v} = \evec{v} \dot{d}(t)
\end{align}

We can see that (\ref{eq:dt1}) will simply lead to

\begin{equation}
    \evec{v}_x =
    \begin{pmatrix}
        1 \\ 0 \\ 0
    \end{pmatrix}, \quad 
    \evec{v}_y =
    \begin{pmatrix}
        0 \\ 1 \\ 0
    \end{pmatrix}, \quad 
    \evec{v}_z =
    \begin{pmatrix}
        0 \\ 0 \\ 1
    \end{pmatrix}, 
\end{equation}
which is also intuitive as this result status that a prismatic joint operating along an axis will only cause a velocity along that axis.

%% file: 6_dETS.tex
\begin{figure}[!t]
    \centering
    \includegraphics[height=6cm]{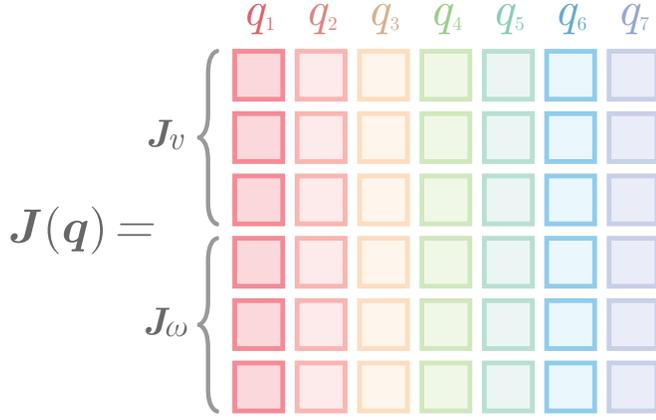}
    \caption{
        Visualization of the Jacobian $\J(\q)$ of the Panda serial-link manipulator. Each column describes how the end-effector pose changes
        due to motion of the corresponding joint.
        The top three rows $\Jv$  correspond to the linear velocity of the end-effector while the bottom three rows $\Jw$ correspond to the angular velocity of the end-effector.
    }
    \label{fig:jacobian}
\end{figure}

Now, we can calculate the derivative of an ETS. To find out how the $j^{th}$ joint affects the end-effector pose,
apply the chain rule to (\ref{m:ets1})

\begin{align} \label{eq:pdr}
    \frac{\partial \matfn{T}{\vec{q}}}
         {\partial q_j} 
    &=
    \frac{\partial} 
         {\partial q_j} \nonumber 
    \left(
        \mat{E}_1(\eta_1) \mat{E}_2(\eta_2) \hdots \mat{E}_M(\eta_M) 
    \right)\\
    &= 
    \prod_{i=1}^{m-1} \mat{E}_i(\eta_i) 
    \frac{\mathrm{d} \mat{E}_m(q_j)} 
         {\mathrm{d} q_j} 
    \prod_{i=m+1}^{M} \mat{E}_i(\eta_i) 
\end{align}
where $\mat{E}_m(q_j)$ is the elementary transform which is a function of the joint coordinate $q_j(t)$. The derivative of the elementary transform with respect to a joint coordinate in (\ref{eq:pdr}) is obtained using one of (\ref{eq:ERx}), (\ref{eq:ERy}), or (\ref{eq:ERz}) for a revolute joint, or one of (\ref{eq:ETx}), (\ref{eq:ETy}), or (\ref{eq:ETz}) for a prismatic joint.

\begin{figure}[!t]
    \centering
    \includegraphics[height=2.1cm]{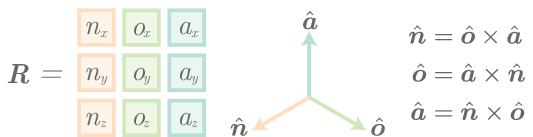}
    \caption{
        The two vector representation of a rotation matrix $\R \in \SO{3}$. The rotation matrix $\R$ describes the coordinate frame in terms of three orthogonal vectors $\evec{n}$, $\evec{o}$, and $\evec{a}$. Each of these three vectors describes the rotation from the reference coordinate frame vectors $\evec{x}$, $\evec{y}$, and $\evec{z}$. As shown above, each of the vectors $\evec{n}$, $\evec{o}$, and $\evec{a}$ can be calculated using the cross product of the other two.
    }
    % \vspace{-0.5cm}
    \label{fig:rotation}
\end{figure}

Combining (\ref{m:dw1}) with (\ref{eq:pdr}), the angular velocity component of the $j^{th}$ column of the manipulator Jacobian is

\begin{align} \label{eq:jwj1}
    \Jw{_{j}}(\q) 
    &=
    \iskx{
        \rho
        \left(
            \frac{\partial \matfn{T}{\vec{q}}}
                {\partial q_j} 
        \right)
        \rho
        \left(
            \matfn{T}{\vec{q}}
        \right)^\top
    }
\end{align}
and combining (\ref{eq:dt1}) with (\ref{eq:pdr}), the translational velocity component of the $j^{th}$ column of the manipulator Jacobian is

\begin{align} \label{eq:jvj1}
    \Jv{_{j}}(\q) 
    &=
    \tau
    \left(
        \frac{\partial \matfn{T}{\vec{q}}}
            {\partial q_j} 
    \right).
\end{align}

Stacking the translational and angular velocity components, the $j^{th}$ column of the manipulator Jacobian becomes 

\begin{equation}
     \J_j(\q) = 
     \begin{pmatrix} 
          \Jv{_{j}}(\q) \\ 
          \Jw{_{j}}(\q) 
     \end{pmatrix} \in \mathbb{R}^{6}.
\end{equation}
where the full manipulator Jacobian as shown in Figure \ref{fig:jacobian} is 

\begin{equation}
    \J(\q) = 
    \begin{pmatrix} 
        \J_1(\q) & \cdots & \J_n(\q)
    \end{pmatrix} \in \mathbb{R}^{6 \times n}.
\end{equation}

\section{Simplifying the Manipulator Jacobian}

We can calculate the manipulator Jacobian using (\ref{eq:jwj1}) and (\ref{eq:jvj1}), however these are computationally expensive.

Expanding (\ref{eq:jwj1}) using (\ref{eq:eta1}) and (\ref{eq:pdr}) and simplify using $\R \R^\top = \mat{I}$ provides

\begin{align} \label{eq:jwj2}
    \Jw{_{j}}(\q) 
    &=
    \iskx{
        \rho
        \left(
            ^0\xi_j
        \right)
        \rho
        \left(
            \sk{\evec{s}_m}
        \right)
        \left(
            \rho(^0\xi_j)^\top
        \right)
    }
\end{align}
where $^0\xi_j$ represents the transform from joint $j$ to the base frame as described by (\ref{m:joint_pose}), and $\sk{\evec{s}_m}$ corresponds to one of the 6 augmented skew symmetric matrices from equations (\ref{eq:ERx})-(\ref{eq:ERz}) and (\ref{eq:ETx})-(\ref{eq:ETz}). 

In the case of a prismatic joint, $\rho(\sk{\evec{s}_m})$ will be a $3 \times 3$ matrix of zeros which will cause the angular velocities to equal 0. In the case of a revolute joint, the angular velocities occur about the axis of rotation in which the joint is oriented, which is poven by (\ref{eq:jwj2})

\begin{align}
    \Jw{_{j}}(\q) 
    &=
    \left\{ 
    \begin{matrix*}[l]
        \evec{o}_j \times \evec{a}_j = \evec{n}_j & \mbox{if} \ \ \mat{E}_m(q_j) = \TR{x}(q_j)\\
        \evec{a}_j \times \evec{n}_j = \evec{o}_j & \mbox{if} \ \ \mat{E}_m(q_j) = \TR{y}(q_j)\\
        \evec{n}_j \times \evec{o}_j = \evec{a}_j & \mbox{if} \ \ \mat{E}_m(q_j) = \TR{z}(q_j)\\
    \begin{pmatrix} 0 & 0 & 0 \end{pmatrix}^\top & \mbox{if} \ \ \mat{E}_m(q_j) = \Tt{}(q_j)\\
    \end{matrix*}
    \right.
\end{align}
where $\rho(^0\xi_j) = \begin{pmatrix} \evec{n}_j & \evec{o}_j & \evec{a}_j \end{pmatrix}$ as described by the two axis convention in Figure \ref{fig:rotation}.

Expanding (\ref{eq:jvj1}) using (\ref{eq:pdr}) provides

\begin{align} \label{eq:jwj2}
    \Jv{_{j}}(\q) 
    &=
        \tau
        \left(
            ^0\xi_j
            \sk{\evec{s}_m}
            {^j\xi}_{e}
        \right)
\end{align}

which reduces to 

\begin{align}
    \Jv{_{j}}(\q) 
    &=
    \left\{ 
    \begin{matrix*}[l]
        \evec{a}_j y_e - \evec{o}_j z_e & \mbox{if} \ \ \mat{E}_m(q_j) = \TR{x}(q_j)\\
        \evec{n}_j z_e - \evec{a}_j x_e & \mbox{if} \ \ \mat{E}_m(q_j) = \TR{y}(q_j)\\
        \evec{o}_j x_e - \evec{y}_j y_e & \mbox{if} \ \ \mat{E}_m(q_j) = \TR{z}(q_j)\\
        \evec{n}_j                      & \mbox{if} \ \ \mat{E}_m(q_j) = \Tt{x}(q_j)\\
        \evec{o}_j                      & \mbox{if} \ \ \mat{E}_m(q_j) = \Tt{y}(q_j)\\
        \evec{a}_j                      & \mbox{if} \ \ \mat{E}_m(q_j) = \Tt{z}(q_j)\\
    \end{matrix*}
    \right.
\end{align}
where $\rho(^0\xi_j) = \begin{pmatrix} \evec{n}_j & \evec{o}_j & \evec{a}_j \end{pmatrix}$ as described by the two axis convention in Figure \ref{fig:rotation}, and $\begin{pmatrix}x_e & y_e & z_e \end{pmatrix}^\top$ correspond to the translational component of ${^j\xi}_{e}$.

%% file: 7_ddT.tex
The second derivative of pose with respect to time is obtained by taking the derivative of \eq{eq:Tdot} by applying the chain rule, product rule, and then chain rule

\begin{align} \label{eq:Tdotdot}
\Tdd &=  
    \frac{\mathrm{d}^2 \mat{T}}
         {\mathrm{d}t^2} \nonumber \\
    &=  
    \sum_{i=1}^{n} 
    \frac{\mathrm{d}}
         {\mathrm{d}t} 
    \left(
        \frac{\partial \T}
            {\partial q_i} 
        \dot{q}_i 
    \right)  \nonumber \\
    &=  
    \sum_{j=1}^{n} 
    \sum_{i=1}^{n}
        \frac{\partial^2 \T}
            {\partial q_j \partial q_i} 
        \dot{q}_i \dot{q}_j
    +
    \sum_{i=1}^{n}
    \frac{\partial \T}
        {\partial q_i} 
    \ddot{q}_i 
    \nonumber \\ 
\end{align}
where $\Tdd \in \mathbb{R}^{4 \times 4}$ and each $\frac{\partial \T}{\partial q_i}, \frac{\partial^2 \T}{\partial q_j \partial q_i} \in \mathbb{R}^{4 \times 4}$.

\begin{figure*}[!t]
    \centering
    \includegraphics[height=5cm]{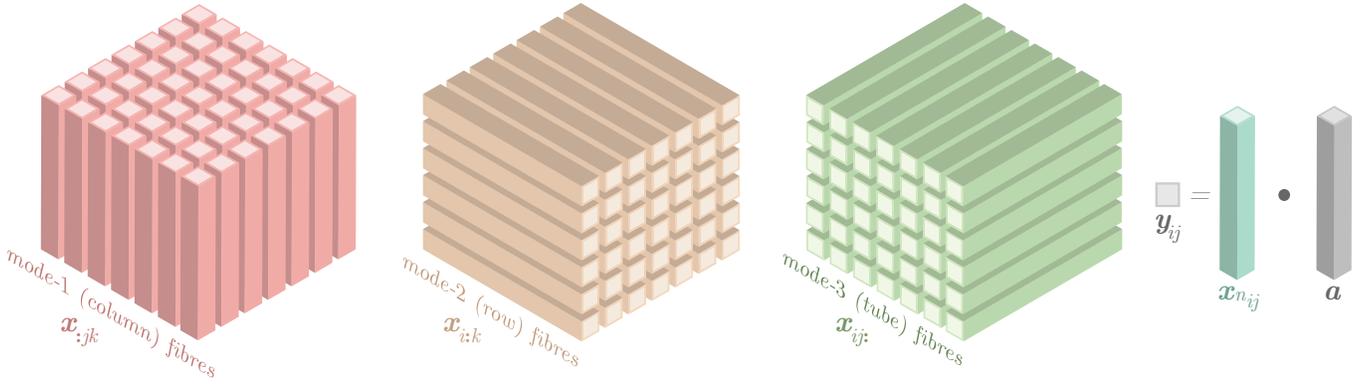}
    \caption{
        Visualization of the fibres (vectors) within a rank-3 tensor $\mat{x} \in \mathbb{R}^{6 \times 7 \times 7}$. The $n$-mode product, denoted by $\bar{\times}_n$, between $\mat{x}$ and the vector $\mat{a} \in \mathbb{R}^7$ results in a rank-2 tensor $\mat{y}$ (a matrix). Each element in the resulting matrix $\mat{y}_{ij}$ is the result of taking the dot product between the mode-$n$ fibres and $\vec{a}$. In general, to take the $n$-mode product, the dimensions of the tensor in mode-$n$ and the vector must be equal \cite{ten}.
    }
    \vspace{-0.5cm}
    \label{fig:product}
\end{figure*}

The information in $\Tdd$ is non-minimal, and redundant, and we can write this as

\begin{equation}
\Tdd = 
    \begin{pmatrix}
        \Rdd & \tdd \\ 
         0   &  0
    \end{pmatrix}
\end{equation}
where $\Rdd \in \mathbb{R}^{3 \times 3}$ and $\tdd \in \mathbb{R}^3$.

We will write the partial derivatives in partitioned form as

\begin{equation}
    \frac{\partial^2 \T}
    {\partial \vec{q} \partial q_i} 
    = 
    \begin{pmatrix} 
    \HR_{i} & \Ht_{i} \\ 
        0  &  0 
    \end{pmatrix} , \,\, 
    \frac{\partial^2 \T}
         {\partial q_j \partial q_i} 
    = 
    \begin{pmatrix} 
        \HR_{ij} & \Ht_{ij} \\ 
         0  &  0 
    \end{pmatrix}
\end{equation}
where the elements can be expressed as 3-way tensors 

\begin{align}
    \HR_{i}
    &=  
    \begin{pmatrix} 
        \HR_{i1} & \cdots & \HR_{in}
    \end{pmatrix} \in \mathbb{R}^{3\times 3 \times n}, \\
    \Ht_{i}
    &=  
    \begin{pmatrix} 
        \Ht_{i1} & \cdots & \Ht_{in}
    \end{pmatrix} \Ht_{i} \in \mathbb{R}^{3\times 1 \times n}
\end{align}
where $ \HR_{ij} \in \mathbb{R}^{3 \times 3}$ and $\Ht_{ij} \in \mathbb{R}^{3 \times 1}$. From this, we can rewrite \eq{eq:Tdotdot} as 

\begin{align}
    &
    \begin{pmatrix} 
        \Rdd & \tdd \\ 
        0   &  0 
    \end{pmatrix} 
    =  \\
    & \quad
    \sum_{i=1}^{n}
    \left(
        \begin{pmatrix} 
            \HR_{i} \bar{\times}_3 \dvec{q} & \Ht_{i} \bar{\times}_3 \dvec{q} \\ 
            0    &  0 
        \end{pmatrix} \dot{q}_i +
        \begin{pmatrix} 
            \JR_{i} & \Jt_{i} \\ 
            0    &  0 
        \end{pmatrix} \ddot{q}_i
    \right)
\end{align}
where $\bar{\times}_n$ represents the $n$-mode product \cite{ten} as explained in Figure \ref{fig:product}.

We can now write a matrix equation for each non-zero partition

\begin{align}
\Rdd &= 
\sum_{i=1}^{n}
\left(
    \HR_{i} \bar{\times}_3 \dvec{q}
\right)  \dot{q}_i + \JR_{i} \ddot{q}_i \label{eq:Rddot} \\
\tdd &= 
\sum_{i=1}^{n}
\left(
    \Ht_{i} \bar{\times}_3 \dvec{q}
\right)  \dot{q}_i + \JR_{i} \ddot{q}_i \label{eq:tddot}
\end{align}

Taking the second equation first, we can simply write

\begin{align} \label{eq:tdd}
    \tdd 
    &= 
    \left(
        \begin{pmatrix} 
            \Ht_{1} & \cdots  & \Ht_{n}  
        \end{pmatrix} 
        \bar{\times}_3
        \qd
    \right)
    \qd +
    \begin{pmatrix} 
        \Jt_{1} & \cdots  & \Jt_{n} 
    \end{pmatrix} 
    \qdd \nonumber \\
    &= 
    \left(
        \Hv \bar{\times}_3 \qd 
    \right)\qd + \Jv \qdd
\end{align}
where $\Hv \in \mathbb{R}^{3 \times n \times n}$.

Rotational acceleration is more complex, using the identity 

\begin{align}
    \skx{\vec{\omega}} = \Rd \R^\top
\end{align}
we can take the derivative using the product rule

\begin{align}
    \skx{\dvec{\omega}} = \Rdd \R^\top + \Rd \Rd^\top.
\end{align}
which we can expand using (\ref{eq:Rdot}) and (\ref{eq:Rddot}) as

\begin{align} \label{eq:skxr2}
    &\skx{\dvec{\omega}} = 
    \sum_{j=1}^{n}
    \sum_{i=1}^{n}
    \HR_{ij} \dot{q}_i \dot{q}_j
    \R^\top  + 
    \sum_{i=1}^{n}
    \JR_{i} \ddot{q}_i \R^\top + \nonumber \\
    & \qquad \qquad
    \sum_{i=1}^{n}
    \JR_i  \dot{q}_i
    \sum_{j=1}^{n}
    (\JR_j  \dot{q}_j)^\top
    \nonumber \\
    & \qquad =
    % \nonumber \\
    % & \qquad
    \sum_{j=1}^{n}
    \sum_{i=1}^{n}
    \HR_{ij} \dot{q}_i \dot{q}_j
    \R^\top  + 
    \sum_{i=1}^{n}
    \JR_{i} \ddot{q}_i \R^\top +  \nonumber \\
    & \qquad \qquad
    \sum_{j=1}^{n}
    \sum_{i=1}^{n}
    \JR_i \JR_j^\top \dot{q}_i \dot{q}_j
    \nonumber \\
    & \qquad = 
    % \nonumber \\
    % & \qquad
    \sum_{j=1}^{n}
    \sum_{i=1}^{n}
    \left(
        \HR_{ij} \R^\top +
        \JR_i \JR_j^\top 
    \right)
    \dot{q}_i \dot{q}_j +  \nonumber \\
    & \qquad \qquad
    \sum_{i=1}^{n}
    \JR_{i} \R^\top \ddot{q}_i 
\end{align}
where each of the terms $\HR_{ij} \R^T + \JR_i \JR_j^\top$, and $\JR_i \R^\top$ must also be skew-symmetric.  This $3\times 3$ matrix equation therefore
has only 3 unique equations so applying the inverse skew operator to both sides we have

\begin{figure*}[b]
    \centering
    \includegraphics[height=5.5cm]{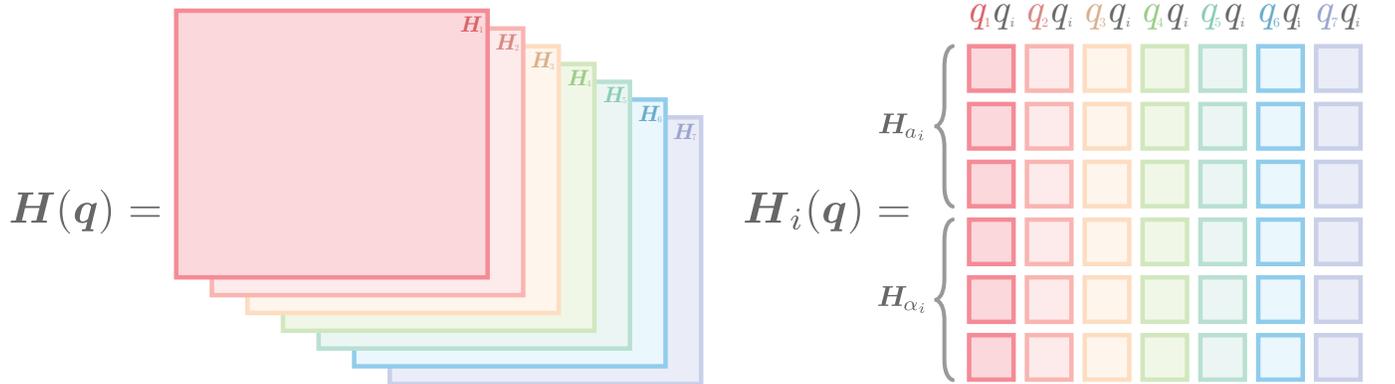}
    \caption{
        Visualization of the Hessian $\H(\q)$ representing the Panda serial-link manipulator. Each slice of the Hessian $\H_i(\q)$ represents the acceleration of the end-effector caused by the velocities of each joint $\q$ with respect to the velocity of joint $q_i$. Within a slice, the top three rows $\H_{\alpha_i}$ correspond to the linear acceleration, while the bottom three rows $\H_{a_i}$ correspond to the angular acceleration, of the end-effector $\vec{\alpha}$ casued by the velocities of two different joints.
    }
    % \vspace{-0.5cm}
    \label{fig:hessian}
\end{figure*}

\begin{align} \label{eq:Rdd}
    \dvec{\omega}  
    &= 
    \sum_{j=1}^{n}
    \sum_{i=1}^{n}
    \iskx{
        \HR_{ij} \R^\top +
        \JR_i \JR_j^\top 
    }
    \dot{q}_i \dot{q}_j + \nonumber \\
    &\phantom{=} \quad
    \sum_{i=1}^{n}
    \iskx{
        \JR_{i} \R^\top
    }
    \ddot{q}_i
    \nonumber \\
    &=
    \sum_{j=1}^{n}
    \sum_{i=1}^{n}
    \Hwx{ij}
    \dot{q}_i \dot{q}_j + 
    \sum_{i=1}^{n}
    \Jw_i \ddot{q}_i \nonumber \\
    &=
    \sum_{j=1}^{n}
    \begin{pmatrix}
        \Hwx{1j} & \cdots & \Hwx{nj}
    \end{pmatrix}
    \qd \dot{q}_j + 
    \begin{pmatrix}
        \Jw_n & \cdots & \Jw_n
    \end{pmatrix}
    \qdd \nonumber \\
    &=
    \left(
        \begin{pmatrix}
            \Hwx{1} & \cdots & \Hwx{n}
        \end{pmatrix}
        \bar{\times}_3
        \qd
    \right)
    \qd + 
    \begin{pmatrix}
        \Jw_n & \cdots & \Jw_n
    \end{pmatrix}
    \qdd \nonumber \\
    &= \left(\Hw(\q) \bar{\times}_3 \qd \right) \qd  + \Jw(\q) \qdd
\end{align}
where $\Hw(\q)  \in \mathbb{R}^{3 \times n \times n}$, $\Hwx{i}(\q)  \in \mathbb{R}^{3 \times n}$, and $\Hwx{ij}(\q)  \in \mathbb{R}^{3}$.

Combining \eq{eq:tdd} and \eq{eq:Rdd} while denoting $\vec{\alpha} = \dvec{\w}$, we can write

\begin{equation} \label{eq:hessian}
    \begin{pmatrix} 
        \vec{a} \\ 
        \vec{\alpha} 
    \end{pmatrix} = 
    \left(
        \begin{pmatrix} 
            \Hv(\q) \\ 
            \Hw(\q) 
        \end{pmatrix} 
        \bar{\times}_3 \qd 
    \right) 
    \qd +
    \begin{pmatrix} 
    \Jv(\q) \\ 
    \Jw(\q) 
\end{pmatrix} \qdd
\end{equation}
which is the second time derivative of the manipulator forward kinematics and

\begin{equation} \label{eq:HT}
    \H(\q) =  
    \begin{pmatrix} 
        \Hv(\q) \\ 
        \Hw(\q) 
    \end{pmatrix} \in \mathbb{R}^{6\times n \times n}
\end{equation}
is the manipulator Hessian tensor expressed in the world-coordinate frame as displayed in Figure \ref{fig:hessian}. More compactly we write

\begin{equation} \label{eq:h0}
     \begin{pmatrix} 
        \vec{a} \\ 
        \vec{\alpha} 
     \end{pmatrix}
     = 
     \left( 
         \H(\q) \bar{\times}_3 \qd 
     \right)
     \qd + \J(\q) \qdd.
\end{equation}
which provides the second derivative of the left side of (\ref{m:ets1}).

%% file: 8_ddE.tex
%%%%%% Column equation -----------------------------------------------------------------------------------------------------------------
\newcounter{mytempeqncnt}

\begin{figure*}[!b] 
% ensure that we have normalsize text 
\normalsize 
% Store the current equation number. 
\setcounter{mytempeqncnt}{\value{equation}} 
% Set the equation number to one less than the one 
% desired for the first equation here. 
% The value here will have to changed if equations 
% are added or removed prior to the place these 
% equations are referenced in the main text. 
\setcounter{equation}{58} 
% IEEE uses as a separator 

\vspace*{4pt} 
\hrulefill % The spacer can be tweaked to stop underfull vboxes. 
\begin{align} \label{eq:ddets}
     \frac{\partial^2 \T}
          {\partial q_k q_j} &= 
     \begin{pmatrix} 
         \HR_{j_k} & \Ht_{j_k} \\ 
          0  & 0 
     \end{pmatrix} \nonumber \\
     &= 
     \frac{\partial} 
          {\partial q_k} 
     \left( 
     \prod_{i=1}^{m-1} \mat{E}_i(\eta_i) 
     \frac{\mathrm{d} \mat{E}_m(q_j)} 
          {d q_j} 
     \prod_{i=m+1}^{M} \mat{E}_i(\eta_i)  
     \right) \nonumber  \\
     &= \left\{ 
     \begin{matrix}
          \prod\limits_{i=1}^{m_k - 1} \mat{E}_i(\eta_i) 
               \frac{\mathrm{d} \mat{E}_{m_k}(\eta_{m_k})}
                    {d q_k}
          \prod\limits_{i=m_k+1}^{m_j - 1} \mat{E}_i(\eta_i) 
               \frac{\mathrm{d} \mat{E}_{m_j}(\eta_{m_j})}
                    {d q_j} 
          \prod\limits_{i=m_j+1}^{M} \mat{E}_i(\eta_i)
          & \mbox{if} \ k < j \\
          \prod\limits_{i=1}^{m_k - 1} E_i(\eta_i) 
               \frac{\mathrm{d}^2 \mat{E}_{m_k}(\eta_{m_k})}
                    {d q_k^2} 
          \prod\limits_{i=m_k+1}^{M} E_i(\eta_i) 
          & \mbox{if} \ k = j \\
          \prod\limits_{i=1}^{m_j - 1} \mat{E}_i(\eta_i) 
               \frac{\mathrm{d} \mat{E}_{m_j}(\eta_{k_2})}
                    {d q_j}
          \prod\limits_{i=m_j+1}^{m_k - 1} \mat{E}_i(\eta_i) 
               \frac{\mathrm{d} \mat{E}_{m_k}(\eta_{m_k})}
                    {d q_k} 
          \prod\limits_{i=m_k+1}^{M} \mat{E}_i(\eta_i)
          & \mbox{if} \ k > j \\
     \end{matrix}
     \right.
\end{align}
% Restore the current equation number. 
\setcounter{equation}{\value{mytempeqncnt}} 
\end{figure*}
% \addtocounter{equation}{2}
%%%%%% Column equation -----------------------------------------------------------------------------------------------------------------

We will now consider the second derivative of a single Elementary Transform. \\

\paragraph{Second Derivative of a Rotation}

We can take the second derivative of a rotation with respect to the rotation angle, by taking the derivative of (\ref{eq:dr1})

\begin{align} \label{eq:ddr0}
    \dfrac{\mathrm{d}}
          {\mathrm{d} \theta} 
    \left(
        \frac{\mathrm{d}\mat{R}(\theta)}
             {\mathrm{d} \theta} 
    \right)
    &=
    \skx{\evec{\omega}} \skx{\evec{\omega}} \ \textrm{e}^{\skx{\evec{\omega}} \theta(t)} \nonumber \\
    \frac{\mathrm{d}^2 \mat{R}(\theta)}
    {\mathrm{d} \theta \mathrm{d} \theta} 
    &=
    \skx{\evec{\omega}}^2 \R(\theta)
\end{align}
and we can also take the derivative of the rotation axis $\evec{\omega}$ in (\ref{m:dw1}) with respect the the rotation angle

\begin{align} \label{m:ddw1} 
    \dfrac{\mathrm{d} \skx{\evec{\omega}}}
          {\mathrm{d} \theta} 
    &=
    \frac{\mathrm{d}^2 \matfn{R}{\theta}}
         {\mathrm{d} \theta^2}
    \matfn{R}{\theta}^\top +
    \frac{\mathrm{d} \matfn{R}{\theta}}
         {\mathrm{d} \theta}
    \left(
        \frac{\mathrm{d} \matfn{R}{\theta}}
             {\mathrm{d} \theta}
    \right)^\top
\end{align}
which we can simplify using (\ref{eq:dr1}) and (\ref{eq:ddr0})

\begin{align} \label{m:ddw2} 
    \dfrac{\mathrm{d} \skx{\evec{\omega}}}
          {\mathrm{d} \theta} 
    &=
    \skx{\evec{\omega}}^2 \R(\theta)
    \matfn{R}{\theta}^\top +
    \skx{\evec{\omega}} \R(\theta)
    \left(
        \skx{\evec{\omega}} \R(\theta)
    \right)^\top \nonumber \\
    &=
    \skx{\evec{\omega}}^2 \R(\theta)
    \matfn{R}{\theta}^\top +
    \skx{\evec{\omega}} \R(\theta)
    \R(\theta)^\top \skx{\evec{\omega}}^\top \nonumber \\
    &=
    \skx{\evec{\omega}}^2 -
    \skx{\evec{\omega}} \skx{\evec{\omega}} \nonumber \\
    &= 0
\end{align}

\paragraph{Second Derivative of a Translation}

We can take the second derivative of a translation with respect to the joint offset by taking the derivative of (\ref{eq:ETx}), (\ref{eq:ETy}) amd (\ref{eq:ETz}). It is clear to see that those matrices are constant and therefore the derivative is equal to zero.

This section may appear confusing however  its implication is simple; a joint within a robot does not experience an acceleration due to the joint's own velocity. However, as we will see in the next section, the acceleration experienced by a joint is influenced by the velocity of the joint's preceding it within the robot.

%% file: 9_ddETS.tex
Now, we can calculate the second derivative of an ETS. By taking the second partial derivative of (\ref{eq:pdr}) with respect to the joint angles we end up with (\ref{eq:ddets}) below, where $(m_k, m_j)$ corresponds to the $i^{th}$ index in (\ref{m:ets1}) in which $q_k, q_j$ respectively appear as a variable. Put simply, in the function $\mat{E}_{m_k}(\eta_{m_k})$, the variable $\eta_{m_k} = q_k$ and in $\mat{E}_{m_j}(\eta_{m_j})$, the variable $\eta_{m_j} = q_j$. 

However, (\ref{eq:ddets}) is rather complex and can be simplified greatly. Lets reconcider the second derivative of a rotation, but with respect to joint angles $q_i$ and $q_j$. We can recalculate (\ref{eq:ddr0}) to be 

\addtocounter{equation}{1}
\begin{align} \label{eq:ddr1}
     \frac{\partial^2 \mat{R}(\q)}
     {\partial q_j \partial q_i} 
     &=
     % \sum_{j=1}^{n}
     % \sum_{i=1}^{n}
     {\skx{\evec{\omega}_j}} {\skx{\evec{\omega}_i}} \R(\q)
\end{align}
and (\ref{m:ddw1}) to be 

\begin{align}
     \dfrac{\partial^2 \skx{\evec{\omega}}}
           {\partial q_j \partial q_i} 
     &=
     \frac{\partial^2 \matfn{R}{\q}}
          {\partial q_j \partial q_i}
     \matfn{R}{\q}^\top +
     \frac{\partial \matfn{R}{\q}}
          {\partial q_i}
     \left(
     \frac{\partial \matfn{R}{\q}}
          {\partial q_j}
     \right)^\top
\end{align}
where we can see from (\ref{eq:skxr2}) that

\begin{align}
     \dfrac{\partial^2 \skx{\evec{\omega}}}
           {\partial q_j \partial q_i} 
     &=
     \skx{\Hwx{ij}}.
\end{align}

We can substitute (\ref{eq:dr1}) and (\ref{eq:ddr1}) into and simplify

\begin{align}
     \Hwx{ij}
     &=
     % \sum_{j=1}^{n}
     % \sum_{i=1}^{n}
     \vee_\times \bigg(
          {\skx{\evec{\omega}_j}} {\skx{\evec{\omega}_i}} \R(\q)
          \matfn{R}{\q}^\top + 
     \bigg. \nonumber \\
     & \qquad
     \bigg.
          {\skx{\evec{\omega}_i}} \R(\q)
          \left(
          {\skx{\evec{\omega}_j}} \R(\q)
          \right)^\top
     \bigg) \nonumber \\
     &=
     % \sum_{j=1}^{n}
     % \sum_{i=1}^{n}
     \iskx{
          {\skx{\evec{\omega}_j}} {\skx{\evec{\omega}_i}} + 
          {\skx{\evec{\omega}_i}} \R(\q) \R(\q)^\top
          {\skx{\evec{\omega}_j}}^\top
     } \nonumber \\
     &=
     % \sum_{j=1}^{n}
     % \sum_{i=1}^{n}
     \iskx{
          {\skx{\evec{\omega}_j}} {\skx{\evec{\omega}_i}} - 
          {\skx{\evec{\omega}_i}}
          {\skx{\evec{\omega}_j}}
     }.
\end{align}

Since we know that $\Jw_x = {\skx{\evec{\omega}_x}}$, and using the identity $\skx{\vec{a}\times \vec{b}} = \skx{\vec{a}}\skx{\vec{b}}-\skx{\vec{b}}\skx{\vec{a}}$ we can show that

\begin{align}
     \Hwx{ij}
     &= \iskx{\skx{\Jw_j} \skx{\Jw_i} - \skx{\Jw_i} \skx{\Jw_j}} \nonumber \\
     &= \Jw_j \times \Jw_i
\end{align}
which means that the rotational component of the manipulator Hessian can be calculated using only the rotational components of the manipulator Jacobian. 
Another relationship is that the velocity of a joint $a$, with respect to the velocity of a preceeding joint $b$, does not contribute acceleration to the end-effector from the perspective of joint $a$. Combining this with the findings of (\ref{m:ddw2}) we can say that $\Hwx{ij}=0$ when $j\geq i$.

For the translational component of the manipulator Hessian $\Hv$, it is easy to show that 

\begin{align} \label{eq:hesst}
     \Hvx{ij}(\q) = 
     \tau
     \left(
     \frac{\partial^2 \T}
          {\partial q_j q_i}
     \right)
\end{align}
where the full expression for $\frac{\partial^2 \T}{\partial q_j q_i}$ is shown in (\ref{eq:ddets}). We can see in (\ref{eq:ddets}) that two of the conditions will have the same result; when $k < j$, and when $k > j$. Therefore, we have

\begin{align}
     \Hvx{ij}(\q) = \Hvx{ji}(\q)
\end{align}
and by exploiting this relationship, we can simplify (\ref{eq:hesst}) to

\begin{align}
     \Hvx{ij}(\q)
     &= \skx{\Jw_a} \Jv{_b} \nonumber \\
     &= \Jw_a \times \Jv{_b}
\end{align}
where $a = \min(i,j)$, and $b = \max(i,j)$. This means that the translational component of the manipulator Hessian can be also be calculated using only the components of the manipulator Jacobian. 

We can now construct the manipulator Hessian. The contribution to end-effector translational and angular acceleration, caused by the velocities of each joint within the robot with respect to the $i^{th}$ joint is 

\begin{align}
     \Hvx{i}(\q)
     &=
     \begin{pmatrix}
         \Hvx{i1}(\q) & \cdots & \Hvx{in}(\q)
     \end{pmatrix} \in \mathbb{R}^{3 \times n} \\
     \Hwx{i}(\q)
     &=
     \begin{pmatrix}
         \Hwx{i1}(\q) & \cdots & \Hwx{in}(\q)
     \end{pmatrix} \in \mathbb{R}^{3 \times n}
\end{align}
which can be stacked into a single matrix 

\begin{align}
     \H_{i}(\q)
     &=
     \begin{pmatrix}
         \Hvx{i}(\q) \\
         \Hwx{i}(\q)
     \end{pmatrix} \in \mathbb{R}^{6 \times n}.
\end{align}

The full manipulator Hessian is obtained by stacking the component Hessians for each joint into a 3-way tensor

\begin{align}
     \H(\q)
     &=
     \begin{pmatrix}
         \H_1(\q) & \cdots & \H_n(\q)
     \end{pmatrix} \in \mathbb{R}^{6 \times n \times n}.
\end{align}

%% file: 9_app.tex
\section{The Special Orthogonal Group} \label{sec:lie}

The study of rotations forms a very important background for this paper. Therefore, we include the following section which has been summarised from \cite{peter}.
\\

\paragraph{Lie Groups}
Rotations in 3-dimensions can be represented by matrices which form Lie groups and which have Lie algebras. These rotations form the Special Orthogonal group in 3 dimensions, also denoted as \SO{3}. Considering the set of all real $3 \times 3$ matrices $\mat{A}$
\begin{align}
    \mat{A} = 
    \begin{pmatrix}
        a_{11} & a_{12} & a_{13} \\
        a_{21} & a_{22} & a_{23} \\
        a_{31} & a_{32} & a_{33} 
    \end{pmatrix}
\end{align}
which could alternatively be written as a linear combination of a set of basis matrices
\begin{align}
    \mat{A} = 
    a_{11}
    \begin{pmatrix}
        1 & 0 & 0 \\
        0 & 0 & 0 \\
        0 & 0 & 0 
    \end{pmatrix} +
    \cdots +
    a_{33}
    \begin{pmatrix}
        0 & 0 & 0 \\
        0 & 0 & 0 \\
        0 & 0 & 1 
    \end{pmatrix}
\end{align}
where each basis matrix represents a direction in a 9-dimensional space of $3 \times 3$ matrices. That is, the nine axes of this space are parallel with each of these basis matrices. Every possible $3 \times 3$ matrix is represented by a point in this space and subsequently any $3 \times 3$ matrix can be represented by a point in this space. However, not all $3 \times 3$ matrices are members of \SO{3} and therefore not all of 9-dimensional space is filled by the group. 

All proper rotation matrices, those belonging to \SO{3}, are a subset of points within the space of all $3 \times 3$ matrices. All points lie in a lower-dimensional subset, a smooth surface, in the 9-dimensional space. This is an instance of a manifold, a lower-dimensional smooth surface embedded within a space. We say that \SO{3} are a matrix Lie group which is closed under the groups operator or composition. For \SO{3}, this is matrix multiplication
\begin{equation*}
    \mat{A} \mat{B} = \mat{C}
\end{equation*}
where $\mat{A}$, $\mat{B}$, and $\mat{C} \in \SO{3}$.
\\

\paragraph{Lie Algebra and the Tangent Space}
Another implication of being a Lie group is that there is a smooth and differentiable manifold structure. At any point on the manifold we can construct tangent vectors. The set of all tangent vectors at that point form a vector space – the tangent space. This is the multidimensional equivalent to a tangent line on a curve, or a tangent plane on a solid. We can think of this as the set of all possible derivatives of the manifold at that point.

The tangent space at the identity is described by the Lie algebra of the group, and the basis directions of the tangent space are called the generators of the group. Points in this tangent space map to elements of the group via the exponential function. If $\vec{g}$ is the Lie algebra for group $\mat{G}$ then
\begin{equation} \label{eq:exp0}
    \forall \mat{X} \in \vec{g} \Rightarrow e^{\mat{X}} \in \mat{G}
\end{equation}
where the elements of $\vec{g}$ and $\mat{G}$ are matrices of the same size and which each have a specific structure. 

The null rotation, represented by the identity matrix, is one point in \SO{3}. At this point we can construct a tangent space which has only 3 dimensions. Every point in the tangent space – the derivatives of the manifold – can be expressed as a linear combination of basis matrices.
\begin{align} \label{eq:gen}
    \mat{\Omega} &=
    \omega_{1}
    \begin{pmatrix}
        0 & 0 & 0 \\
        0 & 0 & -1 \\
        0 & 1 & 0 
    \end{pmatrix} +
    \omega_{2}
    \begin{pmatrix}
        0 & 0 & 1 \\
        0 & 0 & 0 \\
        -1 & 0 & 0 
    \end{pmatrix} + \nonumber \\
    & \qquad \omega_{3}
    \begin{pmatrix}
        0 & -1 & 0 \\
        1 & 0 & 0 \\
        0 & 0 & 0 
    \end{pmatrix} \nonumber \\
    &= \omega_{1}\mat{G}_1 + \omega_{2}\mat{G}_2 + \omega_{3}\mat{G}_3
\end{align}
which is the Lie algebra of the \SO{3} group. The bases of this space: $\mat{G}_1$, $\mat{G}_2$ and $\mat{G}_3$ are called the generators of \SO{3} and belong to \so{3}.
\\

\paragraph{Skew Symmetric Matrices}

A matrix is said to be skew symmetric if 
\begin{equation*}
    \mat{S}^T + \mat{S} = 0.
\end{equation*}

Equation (\ref{eq:gen}) can be written as a skew-symmetric matrix parameterized by the vector $\vec{\omega} = (\omega_1 + \omega_2 +\omega_3)$

\begin{equation} \label{eq:app_dr0}
    \mat{\Omega} = 
    \skx{\vec{\omega}} = 
    \begin{pmatrix}
        0 & -\omega_3 & \omega_2 \\
        \omega_3 & 0 & -\omega_1 \\
        -\omega_2 & \omega_1 & 0 
    \end{pmatrix}
\end{equation}
where the function $\skx{\cdot} : \mathbb{R}^3 \rightarrow \mathbb{R}^{3\times3}$ maps a vector to a skew symmetric matrix. This function reflects the 3 degrees of freedom of the \SO{3} group embedded in the space of all $3\times3$ matrices.

The vex operator $\iskx{\cdot} : \mathbb{R}^{3\times3} \rightarrow \mathbb{R}^3$ is the inverse of $\skx{\cdot}$. For example
\begin{equation} \label{eq:vex}
    \vec{\omega} = \iskx{ \skx{\vec{\omega}} }.
\end{equation}

From (\ref{eq:exp0}) we can say that the exponential of any matrix in \so{3} is a member of \SO{3}
\begin{equation} \label{eq:app_dr1}
    \matfn{R}{\theta, \evec{\omega}} = \textrm{e}^{ \skx{\evec{\omega}} \theta}
\end{equation}
where $\evec{\omega}$ is a unit-vector parallel to the rotation axis, and $\theta$ represents the amount of rotation about that axis.
\\

\paragraph{Derivative of a Rotation}
We can now use the derivative of the manifold in (\ref{eq:app_dr0}) with the rotation matrix in exponential form in (\ref{eq:app_dr1}) to calculate the derivative of a rotation
\begin{align} \label{eq:app_dr2}
    \frac{\mathrm{d} \mat{R}(\theta(t))}
         {\mathrm{d}  \theta(t)} 
    &= \skx{\vec{\omega}} \mat{R}(\theta(t)) 
\end{align}
from which we can obtain the angular velocity of the rotation using (\ref{eq:vex})
\begin{equation}
    \vec{\omega} = \iskx{ \dmat{R} \mat{R}^\top }.
\end{equation}

% \newpage
\section{The Special Euclidean Group} \label{sec:lie2}

Following on form Appendix \ref{sec:lie}, we can introduce the Special Euclidean group in 3 dimensions, also denoted as \SE{3} and commonly referred to as homogenous transformation matrices. Once again, we summarise from \cite{peter}.

These are matrices which also form Lie groups and which have Lie algebras. These are $4 \times 4$ matrices, however, not all matrices in this space are proper homogeneous transformation matrices belonging to \SE{3}. As with rotations, those matrices in \SE{3} lie on a smooth manifold. 

The null motion (zero rotation and translation), which is represented by the identity matrix, is one point in this space. At that point we can construct a tangent space, which has 6 dimensions in this case, and points in the tangent space can be expressed as a linear combination of basis matrices

\begin{align} \label{eq:gen2}
    \mat{\Sigma} = \
    &\omega_{1}
    \begin{pmatrix}
        0 & 0 & 0 & 0 \\
        0 & 0 & -1 & 0 \\
        0 & 1 & 0 & 0 \\
        0 & 0 & 0 & 0 \\
    \end{pmatrix} +
    \omega_{2}
    \begin{pmatrix}
        0 & 0 & 1 & 0 \\
        0 & 0 & 0 & 0 \\
        -1 & 0 & 0 & 0 \\
        0 & 0 & 0 & 0 \\
    \end{pmatrix} + \nonumber \\
    &\omega_{3}
    \begin{pmatrix}
        0 & -1 & 0 & 0 \\
        1 & 0 & 0 & 0 \\
        0 & 0 & 0 & 0 \\
        0 & 0 & 0 & 0 \\
    \end{pmatrix}  + 
    v_{1}
    \begin{pmatrix}
        0 & 0 & 0 & 1 \\
        0 & 0 & 0 & 0 \\
        0 & 0 & 0 & 0 \\
        0 & 0 & 0 & 0 \\
    \end{pmatrix}  + \nonumber \\
    &v_{2}
    \begin{pmatrix}
        0 & 0 & 0 & 0 \\
        0 & 0 & 0 & 1 \\
        0 & 0 & 0 & 0 \\
        0 & 0 & 0 & 0 \\
    \end{pmatrix}  + 
    v_{3}
    \begin{pmatrix}
        0 & 0 & 0 & 0 \\
        0 & 0 & 0 & 0 \\
        0 & 0 & 0 & 1 \\
        0 & 0 & 0 & 0 \\
    \end{pmatrix} 
\end{align}
or in compact form
\begin{equation}
    \mat{\Sigma} = \omega_{1}\mat{G}_1 + \omega_{2}\mat{G}_2 + \omega_{3}\mat{G}_3 + v_1\mat{G}_4 + v_2\mat{G}_5 + v_3\mat{G}_6
\end{equation}
which is the Lie algebra of \se{3}. The bases of this space: $\mat{G}_{1\cdots6}$ are called the generators of \SE{3} and belong to \se{3}.

This can be written in general form as
\begin{equation} \label{eq:gen3}
    \mat{\Sigma} =
    \sk{{\mathcal{S}}} = 
    \begin{pmatrix}
        0 & -\omega_3 & \omega_2 & v_1 \\
        \omega_3 & 0 & -\omega_1 & v_2 \\
        -\omega_2 & \omega_1 & 0 & v_3 \\
        0 & 0 & 0 & 0 \\
    \end{pmatrix}
\end{equation}
which is an augmented skew symmetric matrix parameterized by $\mathcal{S} = (\vec{v,\omega}) \in \mathbb{R}^6$ which is referred to as a twist  and has physical interpretation in terms of a screw axis direction and position and where the function $\sk{\cdot} : \mathbb{R}^6 \rightarrow \mathbb{R}^{4\times4}$ maps a vector to an augmented skew symmetric matrix. The sparse matrix structure and this concise parameterization reflects the 6 degrees of freedom of the \SE{3} group embedded in the space of all $4 \times 4$ matrices.

We extend our earlier vex operator to $\isk{\cdot} : \mathbb{R}^{4\times4} \rightarrow \mathbb{R}^6$. We use this and $\sk{\cdot}$ to convert between a twist representation which is a 6-vector and a Lie algebra representation which is a $4\times4$ augmented skew-symmetric matrix.